\documentclass[runningheads]{llncs}
\usepackage{graphicx}
\usepackage{amsmath}
\usepackage{booktabs} % For formal tables
\usepackage{pifont}
\usepackage{xcolor}
\usepackage{float}
\usepackage[colorlinks = true,
            linkcolor = blue,
            urlcolor  = blue,
            citecolor = blue,
            anchorcolor = blue]{hyperref}
\usepackage{array}
\newcolumntype{P}[1]{>{\centering\arraybackslash}p{#1}}

\usepackage{booktabs} % For formal tables

\usepackage{pifont}

\usepackage{marvosym}

\begin{document}
\title{Conditioning Covert Geo-Location (CGL) Detection on Semantic Class Information}
\author{Binoy Saha\inst{1}\orcidID{0000-0002-8402-2412} \and
Sukhendu Das\inst{2}\orcidID{0000-0002-2823-9211}}
\authorrunning{S. Binoy et al.}
\titlerunning{Conditioning Covert Geo-Location (CGL) Detection on Semantic Class Information}
\institute{
\email{binoysaha@cse.iitm.ac.in}\\ \and
\email{sdas@iitm.ac.in}\\
Visualization and Perception Lab, Dept. of CSE, IIT Madras, India}
\maketitle              % typeset the header of the contribution
\begin{abstract}
The primary goal of artificial intelligence (AI) is to mimic humans. Therefore, to advance toward this goal, the AI community attempts to pick apart certain qualities/skills possessed by humans and tries to imbibe them into machines with the help of datasets/tasks. Over the years, many tasks which require knowledge about the objects present in an image have been proposed and satisfactorily solved by vision models. Recently, with the aim to incorporate knowledge about non-object image regions (hideouts, turns, and other obscured regions), a task for identification of potential hideouts termed Covert Geo-Location (CGL) detection was proposed by Saha et al. It involves identification of image regions which have the potential to either cause an imminent threat or appear as target zones to be accessed for further investigation to identify any occluded objects. Only certain occluding items belonging to certain semantic classes can give rise to CGLs. This fact was overlooked by Saha et al. and no attempts were made to utilize semantic class information, which is crucial for CGL detection. Thus in this paper, we propose a multitask-learning-based approach to achieve two goals - 1) extraction of features having semantic class information; 2) robust training of the common encoder, exploiting large standard annotated datasets as training set for the auxiliary task (semantic segmentation). To explicitly incorporate class information in the features extracted by the encoder, we have further employed attention mechanism in a novel manner. In this work, we have also proposed a better evaluation metric for CGL detection that gives more weightage to recognition rather than precise localization. Experimental evaluations performed on the CGL dataset, demonstrate a significant increase in performance of about 3\% to 14\% mIoU and 3\% to 16\% DaR on split 1 and $\approx$1\% mIoU and 1\% to 2\% DaR on split 2 over SOTA, serving as a testimony to the superiority of our approach.
\\
\textbf{ACK: IMPRINT (MHRD/DRDO) GoI, for support}
\end{abstract}
\section{Introduction}
\begin{figure}[!t]
  \centering
  \includegraphics[scale=0.57]{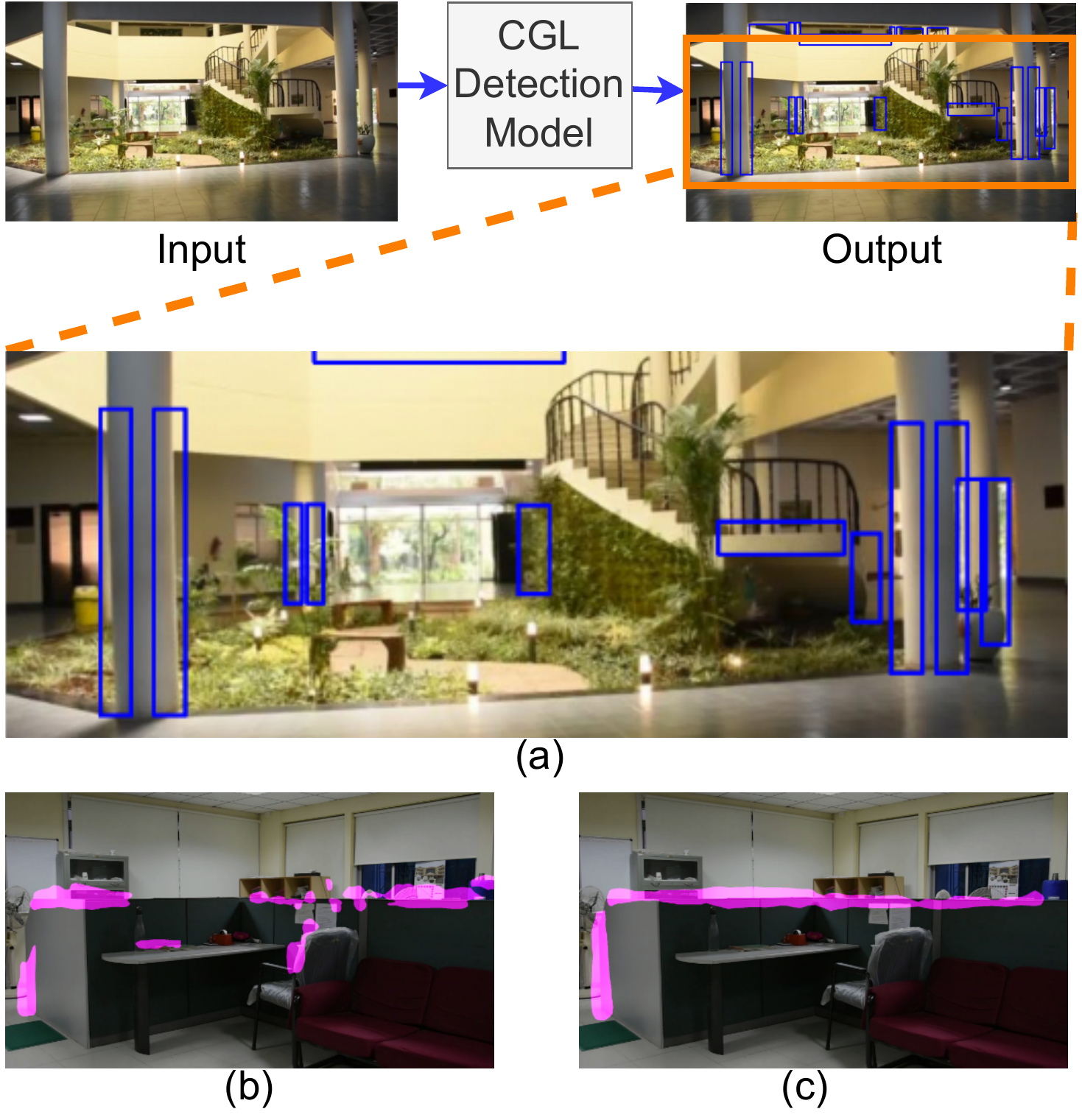}
  \caption{(a) input and the expected output for CGL detection. Blue bounding boxes indicate ground truth CGLs. (b) shows output of the model proposed in \cite{saha2022catch}. (c) shows the output of our model. In (b) and (c), pink-colored translucent mask has been used to indicate predicted CGL blobs in the output of models (segmentation masks). Our model successfully extracts semantic class information. It exhibits better understanding of the objects/items present in the scene and avoids partial detections as observed in (b).   
  }
  \label{intro_fig}
\end{figure}
The majority of vision-related tasks \cite{krizhevsky2012imagenet,lin2014microsoft,Everingham15,krishnavisualgenome,balanced_vqa_v2} entail identifying only the objects present in an image. However, knowledge about the non-object image regions like hiding places, corners, bends, and other obscured areas of the scene can also provide helpful information for a variety of tasks like automated navigation, surveillance, etc. Detecting hidden spots or covert locations in a scene, for instance, could provide a promising next step towards achieving a significant goal of better scene understanding. Thus, in this work, we attempt to tackle a recently proposed task \cite{saha2022catch} termed Covert Geo-Location (CGL) Detection, where, given an input image, the target is to identify and localize potential hideouts (Covert Geo-Locations) in the
image. Figure \ref{intro_fig}(a) shows the input and the expected output in the case of CGL detection.

Covert locations are hidden areas of the scene that are typically not directly visible from the camera viewpoint. Occluding items such as pillars, doors, furniture, etc. can create such covert locations.   An intelligent agent can be deployed to assess the environment to identify potential danger zones around occluding items. The goal in CGL detection is thus to identify certain sub-segments of boundaries of occluding items (like furniture items, pillars, curtains), which need to be accessed for further investigation. These locations may either have the potential to cause an imminent threat or appear as target zones to be accessed for further investigation to identify any occluded objects.

In the absence of any prior work, Saha et al. \cite{saha2022catch} presented a novel dataset for CGL detection consisting of real-world images depicting diverse indoor environments, baseline models, and a detailed analysis of the challenges posed by CGL detection. Moreover, a novel segmentation-based Depth-aware Feature Learning Block (DFLB) was also proposed that facilitated the extraction of relevant depth features (with a single RGB image as input) required for the proposed task. Additionally, two novel feature-level loss functions were also proposed by the authors, namely, Geometric Transformation Equivariance (GTE) loss and Intraclass Variance reduction (IVR) loss to enforce additional constraints on the model to make it recognize the underlying depth pattern in all CGLs. However, in order to successfully detect or segment CGLs, the model needs to understand what kind of objects/occluding-items give rise to CGL and what items do not give rise to a CGL. This aspect was not considered by \cite{saha2022catch} while designing the model.

\begin{figure}[!t]
  \centering
  \includegraphics[scale=0.52]{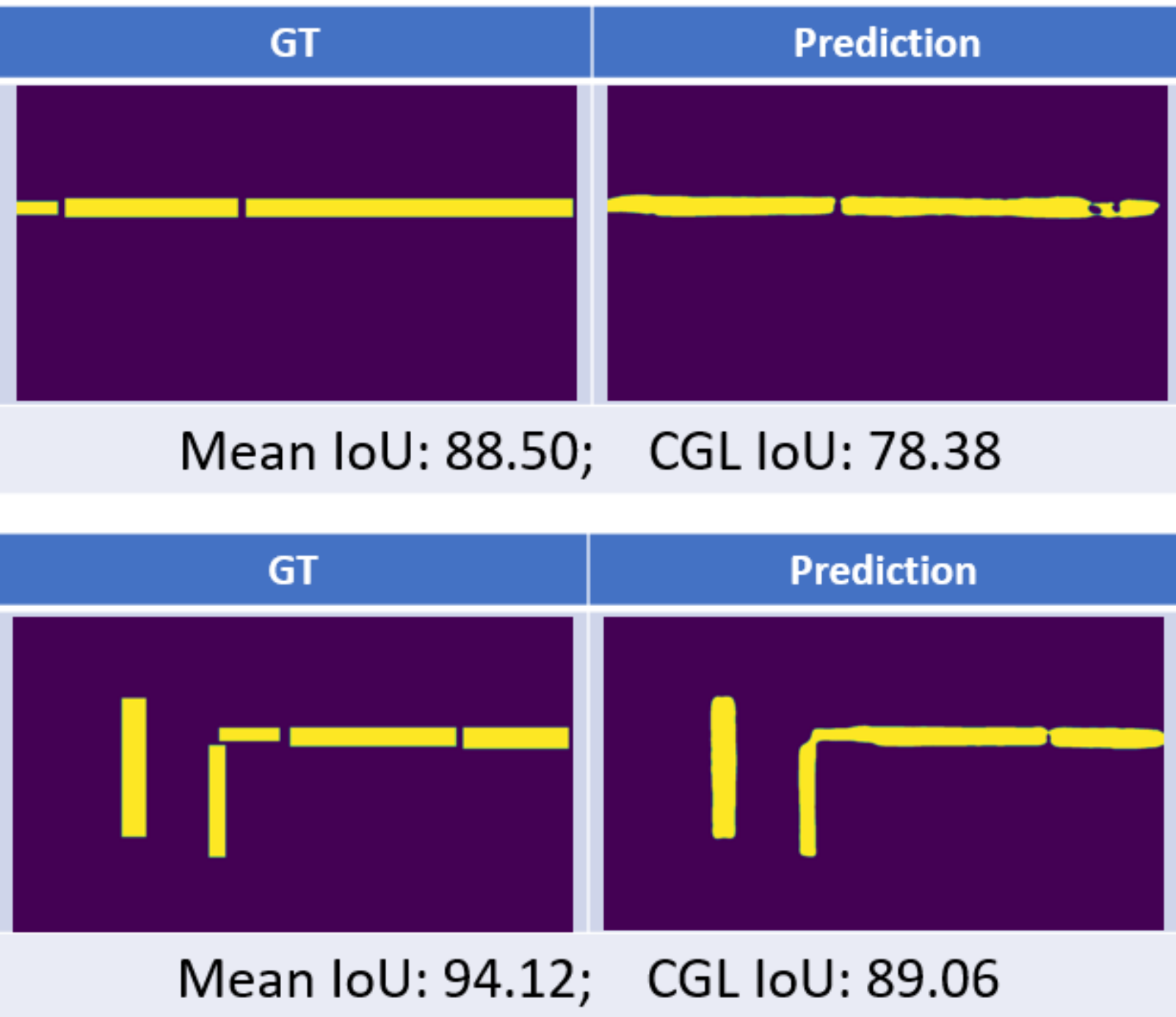}
  \caption{shows GT and predicted segmentation mask for two sample images from the CGL dataset. The proposed model was used to get the segmentation masks. Each row shows GT and prediction for a sample image and the corresponding Mean IoU and CGL IoU scores are mentioned at the bottom of each row. Yellow blobs indicate CGL blobs and the background/non-CGL class has been indicated using purple color.}
  \label{motivation_eval}
\end{figure}
 
In this paper, we thus attempt to condition CGL detection on the semantic class of the occluding items. This information about the semantic class of occluding items can be implicitly extracted using a multitask learning setting. Thus, in this work, we exploit a multitask learning-based approach which also provides a way of leveraging large standard datasets for better training of the feature extractor. This work builds on top of \cite{caruana1997multitask}, which first proposed multitask learning and provided evidence that learnings from one task indeed expedite learning for other tasks. The network is trained using two datasets (proposed CGL dataset and a standard dataset - for our experiments we have used the ADE20k dataset \cite{zhou2017scene,zhou2019semantic}). Input images from both datasets are passed through a common CNN-based feature extractor. Then the feature volume obtained is passed to two different decoders, one of which is trained to perform CGL segmentation and the other is trained to perform semantic segmentation. We have further employed multi-head self/cross-attention-based decoders for explicit propagation of semantic class information to the CGL segmentation decoder. Figure \ref{intro_fig}(b) shows the output of the model proposed in \cite{saha2022catch} and figure \ref{intro_fig}(c) shows the output of the proposed model. Evidently, the proposed model manifests better semantic understanding and outperforms other models.  

The height and width of CGL cannot be defined precisely (in most cases) in the ground truth and thus the annotations inherently contain some uncertainty even though certain protocols were followed by Saha et al. \cite{saha2022catch} during annotation process. Thus, detection of all CGLs is more crucial than detecting CGLs of the same height and width (dimensions) as those in GT. So, ideally, we want an evaluation metric that can give more weightage to recognition than precise localization in the performance score. Mean IoU (which was also used by \cite{saha2022catch} for performance comparison of CGL detection models) is a standard evaluation metric for evaluating segmentation models, but it performs per-pixel evaluation by considering all corresponding image coordinate positions in GT and the output of the model (evident from figure \ref{motivation_eval}), which is not desired for the evaluation of a CGL detection model. To tackle this problem, we have proposed a novel dimension (height and width) agnostic evaluation metric for CGL detection in this paper. More details about the proposed metric are included in section 4.1. 

Experimental evaluations, performed using CGL Dataset, demonstrate a significant increase in performance (improvement is observed in all three metrics - Mean IoU, CGL IoU, and the proposed DaR metric) over the existing segmentation models (when adapted and trained from scratch for CGL Detection), serving as a testimony to the superiority of our proposed approach. Several illustrations have been provided to prove the efficacy and the superiority of the proposed evaluation metric (DaR) over the existing metric.

\vspace{3mm}
\textbf{Our key contributions can be summarized as follows:}
\begin{itemize}
  \item We propose a multitask-learning-based strategy for extracting features with semantic information and robust training of the common feature extractor (using large standard annotated datasets as the training set for an auxiliary task, semantic segmentation in our case.)
  \item To explicitly incorporate object information in the features extracted by the common encoder, a multi-head attention mechanism has been utilized in a novel manner.
  \item We have also proposed a better evaluation metric for CGL detection which prioritizes the identification of CGLs over accurate localization.
  \item Our method beats the model proposed in \cite{saha2022catch} on both train/test splits of the CGL detection dataset. On split 1, we achieve a performance improvement of 3\% to 14\% mIoU and 3\% to 16\% DaR and on split 2, performance improvement is $\approx$1\% mIoU and 1\% to 2\% DaR.
  
\end{itemize}

\section{Related Work}
CGL detection requires context-aware detection and understanding of the complex 3D spatial relationships between edges of occluding items and their surroundings. Depth information is thus very crucial for the detection of CGLs. For this reason, authors of \cite{saha2022catch} developed an approach that can effectively extract RGB-based features as well as relevant depth features, using only a single RGB image as input. A novel DL-based technique was proposed which used an auxiliary decoder block, named Depth-aware Feature Learning Block (DFLB), to steer the common feature extractor towards extraction of necessary depth features (along with other RGB-based features). Additionally, as the proposed dataset (1.5K CGL annotated images) was relatively small, they also leveraged two novel self-supervised feature-level loss functions namely, Geometric Transformation Equivariance (GTE) loss and Intraclass Variance reduction (IVR) loss to enforce additional constraints on the model to make it recognize key aspects of CGLs which are helpful for their detection. 

In this paper, we have used multitask learning and attention mechanism in a novel manner to design a model for CGL detection that considers the semantic class of the items present in the scene to perform the task at hand. In the subsequent subsections (2.1, 2.2, 2.3), we briefly present some technical details about these concepts.

\subsection{Multi-Task Learning (MTL)}

Majority of the time, handling a single task at a time is what we are most concerned with in Machine Learning (ML) scenarios. Using data to accomplish a specific task or to maximize/minimize one metric or objective function at a time is how an ML challenge or task is often framed. This technique eventually reaches a performance limit because of the quantity of the datasets or the model's capacity to extract meaningful representations from them. On the other hand, Multi-Task Learning (MTL) is a machine learning strategy that aims to learn many tasks concurrently while simultaneously minimizing multiple loss terms. In MTL, a single model is expected to learn to perform all of the tasks simultaneously rather than training separate models for each task. The model uses all of the available data from the various datasets  (for various tasks) to learn generalized representations for the data. The simplest approach is to minimize a linear combination of the loss functions of the individual tasks. Each task will have its individual loss function. So in most multi-task learning-based models \cite{zhang2014facial,dai2016instance,zhao2018modulation,liu2019end,ma2018modeling,Misra_2016_CVPR}, each loss function is simply given its own weight and the sum of these weighted losses is minimized.

\[ \min_{\theta} \sum_{i=1}^{T} w_{i}\mathcal{L}_{i}(\theta, D_{i}) 
\]

where, $\theta$ represents parameters of the entire model, $w_{i}$ represents weight assigned to the loss component for $i^{th}$ task, $\mathcal{L}_{i}(\theta, D_{i})$ represents the loss for the $i^{th}$ task, and $D_{i}$ represents the dataset used for the $i^{th}$ task.

% MTL is widely used in a variety of fields, including natural language processing, computer vision, and recommendation systems. It is also widely used in the industry because of its capacity to successfully use vast volumes of data to solve related problems.

In our work, we have used a feature learning-based MTL approach. Caruana et al. \cite{caruana1997multitask} proposed one of the earliest models for multi-task learning. The output layer has "m" output units (one for each task), while the input layer receives training instances from all of the tasks. Here, the hidden layer's output can be treated as the common feature representation learned for the "m" tasks. The distinction between MTL networks and multi-layer feedforward neural networks used for single-task learning is in the output layer, where MTL networks have "m" nodes as opposed to just one output unit as in the case of single-task learning.

\subsection{Multi-head Attention}
Attention has been used in a variety of applications in computer vision, including image classification, image segmentation, and image captioning. Spatial and feature-based attention (channel-wise attention) are two types of attention-based processes. In case of spatial attention \cite{carion2020end,yuan2018ocnet,wang2018non}, different weights are assigned to different spatial locations in spatial attention, but these weights are maintained throughout all feature channels at all spatial locations. \cite{xu2015show} has proposed one of the most essential image captioning systems based on spatial attention. A CNN is used as an encoder in their model, which extracts a number of feature vectors (or annotation vectors), each of which corresponds to a distinct region of the picture, allowing the decoder to focus on specific image sections. In comparison, channel-wise attention permits individual feature maps (channels) to be assigned their own weight/attention values. As an example, the encoder-decoder framework of \cite{chen2017sca} applied to perform image captioning incorporates spatial as well as channel-wise attention in the same CNN.

Attention can be generalized and considered as the weighted sum of the values based on the query and the associated keys, given a set of key-value pairs (K, V) and a query (q). The query "attends" to the values by choosing which keys to focus on.

Multi-head attention uses an attention mechanism several times in parallel with different attention heads having a different set of learnable parameters. The attention outputs from all the attention heads are then concatenated and linearly transformed into the desired dimension. Intuitively, multiple attention heads allow for attending to feature vectors within a feature volume differently (calculating attention scores with different criteria). For example, in the case of image data, a few attention heads could model global spatial context and a few others could model local spatial context. Multi-head attention is given as:

\begin{equation}
  M\_Att = [H_{1},H_{2},...,H_{h}]W_{o}
\end{equation}
where, 
\begin{equation}
  H_{i} = SPDA(QW^{Q}_{i}, KW^{K}_{i}, VW^{V}_{i})
\end{equation}
and Scaled dot-product attention (SPDA) is an attention mechanism where the outputs of dot products are scaled down by $\sqrt{d_{k}}$ as follows,
\begin{equation}
  SPDA(QW^{Q}_{i}, KW^{K}_{i}, VW^{V}_{i}) = softmax(\frac{QW^{Q}_{i}(KW^{K}_{i})^{T}}{\sqrt{d_{k}}})VW^{V}_{i}
\end{equation}

$W^{Q}, W^{K}, W^{V}, W_{0}$ are all learnable parameter matrices and $d_{k}$ is dimensionality of feature vectors.

Note that scaled dot-product attention (SPDA) is the most commonly used mechanism to compute attention scores, although in principle it can be replaced by any other type of attention mechanism like multiplicative attention, additive attention, etc.

\subsection{Multi-head Self-attention/Cross-attention}
The only difference between cross-attention and self-attention lies in the inputs to the two mechanisms. Two different embedding sequences of the same dimension are combined asymmetrically by cross-attention. Self-attention input, on the other hand, consists of just one embedding sequence. In the case of image data, feature volume extracted by the feature extractor is treated as a sequence of feature vectors. In cross-attention, one of the sequences serves as a query, while the other acts the key and value. In self-attention, the same sequence acts as all three components (query, key, value).  

\section{Proposed Approach}

\begin{figure*}[h!]
  \centering
  \hspace*{0.02in}
  \includegraphics[scale=0.22]{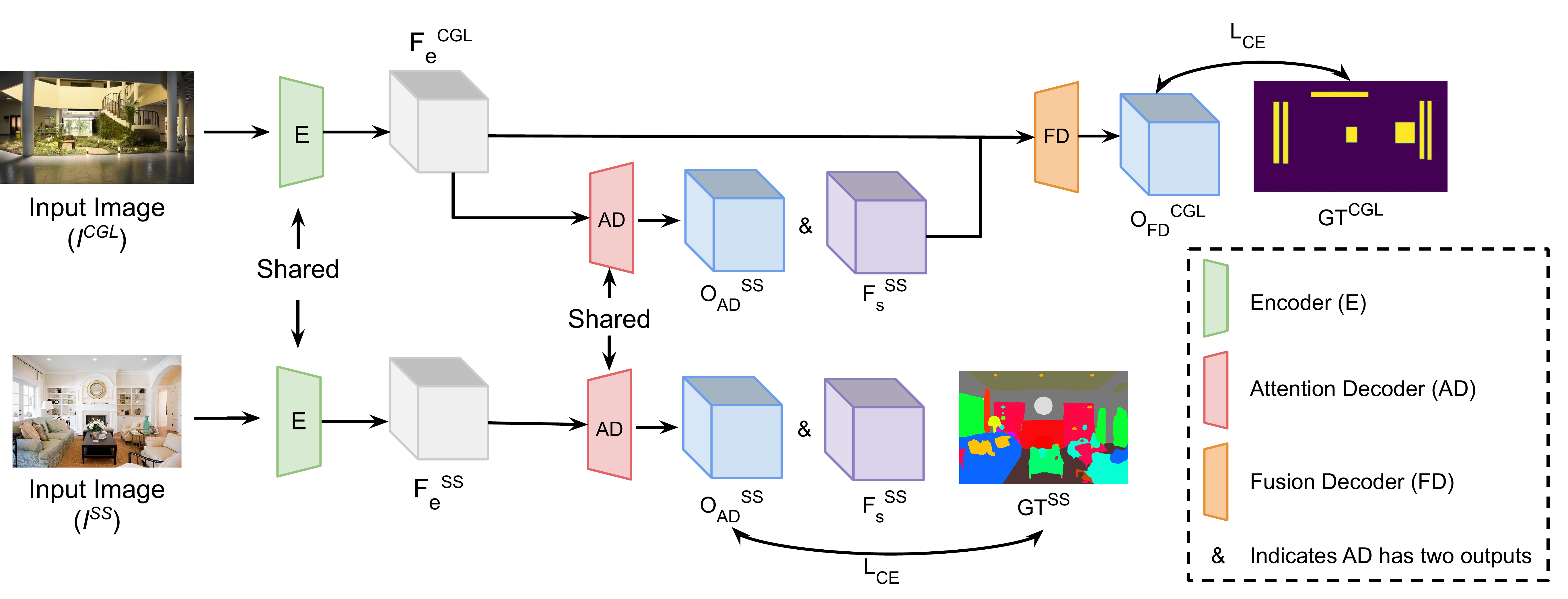}
  \caption{Proposed architecture. Multi-task learning setting has been exploited with multi-head self-attention and cross-attention-based decoders to condition CGL segmentation on semantic class information.
%   The plus ($+$) symbol has been used to indicate that the attention decoder generates two output volumes.
  }
  \label{multitask_architecture}
\end{figure*}

By concentrating on a single task, we can typically attain satisfactory results, but the model still misses out on information that could help it perform better on the metric we care about. Training the model simultaneously on multiple related tasks helps prevent this as the model gets to derive knowledge from multiple training signals coming from the various tasks being used for training. Also, we can improve the generalization of the model on our initial task by sharing representations between similar tasks. This method is known as Multi-Task Learning (MTL), and we have utilized it in this work.

Furthermore, in order to successfully detect or segment CGLs, the model needs to understand what kind of objects/occluding items give rise to CGL and what items cannot give rise to a CGL. This information about the semantic class of occluding items can again be implicitly extracted using a multitask learning setting. The images from CGL dataset can also be then passed through a semantic segmentation decoder to get semantic class information. This semantic class information can be then used by the CGL segmentation branch yielding better performance.  

Thus, to extract features having semantic class information and to exploit larger standard datasets for robust training of the common feature extractor, we have proposed a multitask learning-based architecture. Overall architecture is shown in figure \ref{multitask_architecture}. One of the decoder branches (Fusion Decoder) is trained to perform CGL segmentation (using CGL dataset for supervision) and the other decoder branch (Attention Decoder) is trained to perform semantic segmentation and is trained using the ADE20k dataset \cite{zhou2017scene,zhou2019semantic}. The attention decoder can also generalize and segment the input image from the CGL dataset and thus can provide additional semantic class information, which can further aid the fusion decoder in performing CGL segmentation. 

The output of deeper layers of the decoder contains class-specific activation maps. These activation maps can be exploited for better propagation of semantic class information. We have employed multi-layered multi-head attention to achieve the same. Specifically, we propose to use multi-head self-attention layers to obtain better and more global features in the Attention Decoder (AD) and we propose to use multi-head cross-attention between encoder feature maps and the self-attended feature volume generated by AD in the Fusion Decoder (FD).

\subsection{Notations}
Following Saha et al., we attempt to solve CGL detection as a segmentation task. The target in segmentation is to assign a class label to every pixel of the input image. Formally defined, given an RGB input image $I$ $\in$  $R^{3\;\times\;H\;\times\;W}$, the goal is to generate a classification map of the same size with the values in the map ranging from 0 to K-1, where K represents the number of classes in the dataset used. The number of classes (K) is 2 (CGL and non-CGL/background) in the case of CGL segmentation and 150 in the case of semantic segmentation (since there are 150 classes in the ADE20k dataset). Input image pairs (one from CGL dataset $I^{CGL}$ and one from ADE20k dataset $I^{SS}$) are first passed through the common encoder to get encoder feature maps denoted by $F^{CGL}_{e}$ and $F^{SS}_{e}$ respectively, where $F^{CGL}_{e}$ and $F^{SS}_{e}$ $\in$ $R^{d\;\times\;H/d'\;\times\;W/d'}$. These encoder feature maps are passed to corresponding decoders ($F^{CGL}_{e}$ to attention and $F^{SS}_{e}$ to fusion decoder) that generate label maps $O^{SS}$ and $O^{CGL}$ respectively. where, $O^{SS}$ $\in$ $R^{150\;\times\;H/4\;\times\;W/4}$, and $O^{CGL}$ $\in$ $R^{2\;\times\;H/4\;\times\;W/4}$. We denote the ground truth segmentation masks for attention decoder and fusion decoder by $GT^{SS}$ and $GT^{CGL}$ respectively, where $GT^{SS}$ and $GT^{CGL}$ $\in$ $R^{H/4\;\times\;W/4}$ are binary maps. The output volumes denoted as $O^{SS}_{AD}$ and $F^{SS}_{s}$, are obtained when encoder feature maps for an image in CGL dataset are passed through the attention decoder.

\subsection{Attention Decoder}
\begin{figure}[h!]
  \centering
  \includegraphics[scale=0.28]{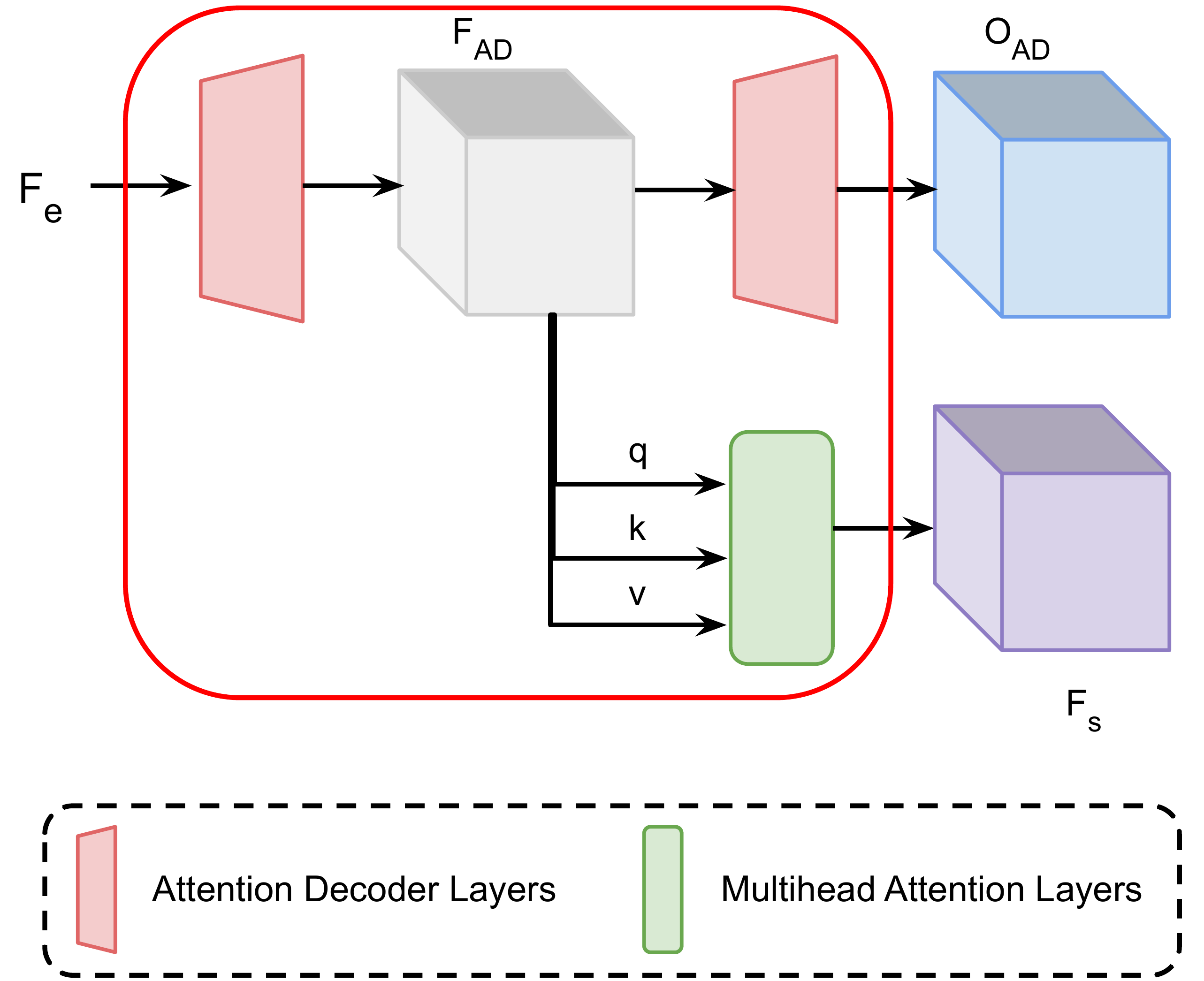}
  \caption{\textbf{Attention Decoder}. The red box encloses the attention decoder. It takes feature volume ($F_{e}$) extracted by the feature extractor as input and outputs a self-attended version of the feature volume ($F_{s}$) and per-pixel semantic classification ($O_{AD}$) as output. It employs multi-layered multi-head self-attention to obtain $F_{s}$ from $F_{AD}$, which is the output of an intermediate layer in the attention decoder. q,k, and v represent the query, key, and value components respectively.}
  \label{attention_decoder}
\end{figure}

CGL detection requires information about the local spatial context. Contextual information can be extracted either by widening the receptive field and/or by using attention mechanism. We have used multi-head self-attention to model global contextual information. \\
The encoder feature maps are passed through a series of decoder layers. Deeper layers of the decoder have class-specific activation maps but they are prone to become heavily dataset-specific. We want our model to generalize well on images having occluding items not seen during training. So we exploit the output of an intermediate decoder layer (say $C_{i}$) to strike a balance between the effective extraction of semantic class information and features that can be generalized. 
\\
We represent the feature map of the intermediate decoder layer $C_{i}$ as $F_{AD}$. The intermediate feature map $F_{AD}$ is then passed through multi-head self-attention layers (as shown in figure \ref{attention_decoder}) to obtain $F_{s}$. The output of the final layer of the decoder is the final classification output ($O_{AD}$). This decoder is used to perform semantic segmentation and it has been trained using the ADE20k dataset \cite{zhou2017scene,zhou2019semantic} for our experiments. Any other semantic segmentation dataset or in fact any other vision dataset could also have been used with the final layer modified according to the annotations in the dataset.

\subsection{Fusion Decoder}

\begin{figure}[h!]
  \centering
  \includegraphics[scale=0.28]{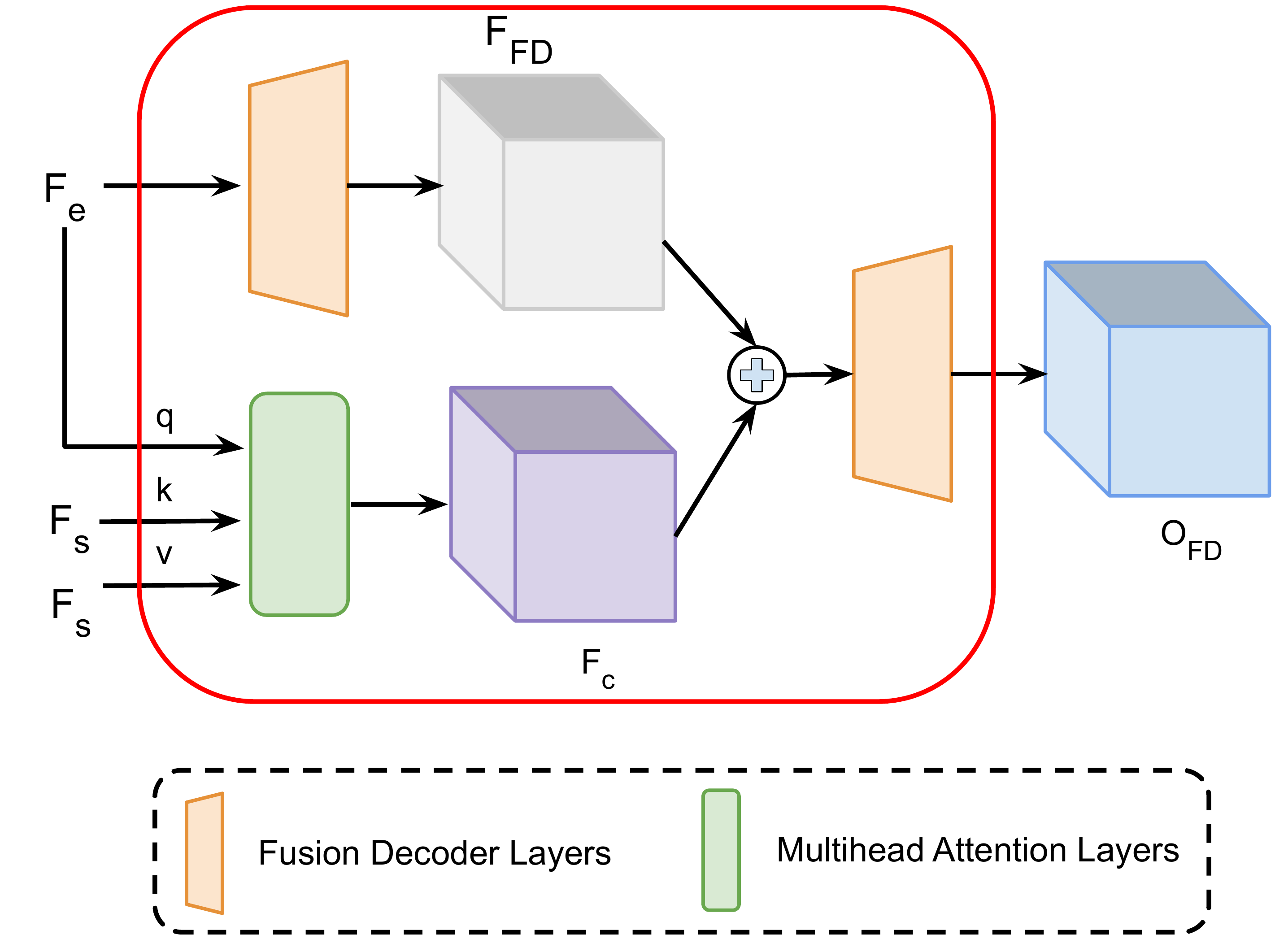}
  \caption{\textbf{Fusion Decoder}. The red box encloses the fusion decoder. It takes encoder feature map ($F_{e}$) and self-attended decoder feature volume ($F_{s}$) from attention decoder as input and outputs CGL segmentation mask ($O_{FD}$). It employs multi-layered multi-head cross-attention with $F_{e}$ as query and $F_{s}$ as key and value, and then fuses the cross-attended feature volume $F_{c}$ with $F_{e}$.}
  \label{fusion_decoder}
\end{figure}

The fusion decoder (overall architecture shown in figure \ref{fusion_decoder}) is used to effectively fuse semantic class information contained in ($F_{s}$) with the encoder feature maps ($F_{e}$). This can be done by using attention mechanism with encoder feature map $F_{e}$ as the query (q) and the self-attended decoder feature maps ($F_{s}$) as key (k) and value (v).

Multi-layered multi-head cross-attention is used to obtain feature map $F_{c}$ having semantic class information. This is then fused with the output of an intermediate layer of fusion decoder using element-wise addition operation. The fused feature map is then finally passed through another series of decoder layers to obtain the final CGL segmentation output $O_{FD}$. This decoder is supposed to perform CGL segmentation and it is supervised using our proposed CGL dataset.

\subsection{Loss functions}
The proposed model is trained using the following loss function,
\begin{equation}
  \mathcal{L} = \alpha * \mathcal{L_{_{CGL}}} + \beta * \mathcal{L_{_{SS}}}
  \label{eq_total_loss_2}
\end{equation}
where, $\alpha$, $\beta$ are hyperparameters. Hyperparameter tuning was performed using the K-fold cross-validation technique. Equations \ref{eq_loss_part1} and \ref{eq_loss_part2} represent $\mathcal{L_{_{CGL}}}$ and $\mathcal{L_{_{SS}}}$ loss terms respectively.   

\begin{equation}
  \mathcal{L_{_{CGL}}} = \mathcal{L_{_{CE}}}(O_{_{FD}}^{CGL}, GT^{CGL})
  \label{eq_loss_part1}
\end{equation}

\begin{equation}
  \mathcal{L_{_{SS}}} = \mathcal{L_{_{CE}}}(O_{_{AD}}^{SS}, GT^{SS})
  \label{eq_loss_part2}
\end{equation}

where, $\mathcal{L_{_{CE}}}$ is standard cross-entropy loss. $GT^{SS}$ and $GT^{CGL}$ represent the ground truth segmentation masks for the attention decoder and the fusion decoder respectively. $O^{CGL}_{FD}$ represents the output of the fusion decoder for an image from the CGL dataset and $O^{SS}_{AD}$ represents the output of the attention decoder from an image from ADE20k \cite{zhou2017scene,zhou2019semantic}.

\section{Results and Experiments}
We have used the ADE20k dataset \cite{zhou2017scene,zhou2019semantic} along with the CGL detection dataset \cite{saha2022catch} for training our model. We evaluate our models for CGL segmentation using the proposed dataset. The CGL detection dataset contains 1500 DSLR captured real-world CGL-centric images of dimensions 1080 $\times$ 1920 (H $\times$ W). Saha et al. proposed two train/test splits: split 1 and split 2.  Split 1 divides images of the dataset such that, if the test partition contains images depicting environments/scenes such as offices, lobbies, etc., the training partition images depict different types of scenes such as classrooms, houses, etc. This ensures that the input scenes are unknown at test time. On the other hand, split 2 allows some overlap between train/test partition in terms of scenes depicted in the images, specifically, a few images ($\approx$17\%) of the test partition scenes are also incorporated in the training partition. We evaluate our models on both train/test splits.

We use Adam optimizer and a similar training scheme to train all models. We report mean IoU, and CGL IoU (IoU for CGL class) for comparison of models. To provide a fair and informative study, we also report performance on the proposed \textbf{D}imension \textbf{a}gnostic \textbf{R}ecall (DaR) metric. 

\subsection{Proposed Evaluation Metric}
\begin{figure*}[h!]
  \centering
  \includegraphics[scale=0.23]{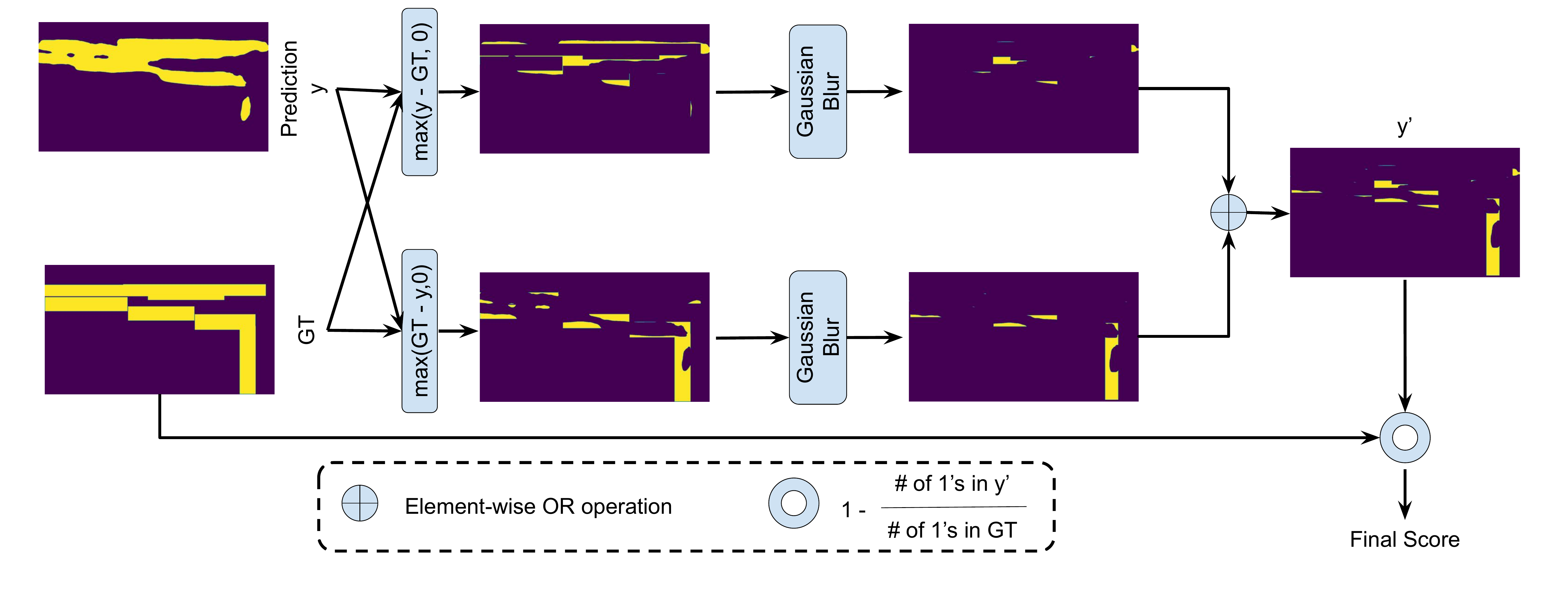}
  \caption{The process used for computing the proposed Dimension-agnostic Recall (DaR) metric score. For the purpose of illustration, the whole process has been shown with the help of an example. The prediction (y) and the corresponding GT have been shown for a sample image from our CGL dataset. Yellow blobs indicate CGLs and purple regions indicate background/non-CGL regions. The white blobs in y' indicate regions of disagreement between prediction (y) and ground truth.}
  \label{new_eval}
\end{figure*}

Figure \ref{new_eval} shows the flowchart of the process used to obtain the score for the proposed metric named "Dimension-agnostic Recall" (DaR). At first, we obtain two relative complements, specifically, "$y - GT$" and "$GT - y$". Where "$y - GT$" represents false positives, "$GT - y$" represents false negatives, and "$-$" represents element-wise subtraction operation. Subsequently, to overlook minor mismatch in the dimensions of predicted CGLs, we make use of a symmetric Gaussian blur kernel with two parameters ($\sigma$, $Th$), where $\sigma$ represents the standard deviation of the Gaussian kernel and $Th$ is a threshold. We have empirically set $\sigma$ to 3.0 and $Th$ to 0.999.

\begin{figure}[h!]
  \centering
  \includegraphics[scale=0.5]{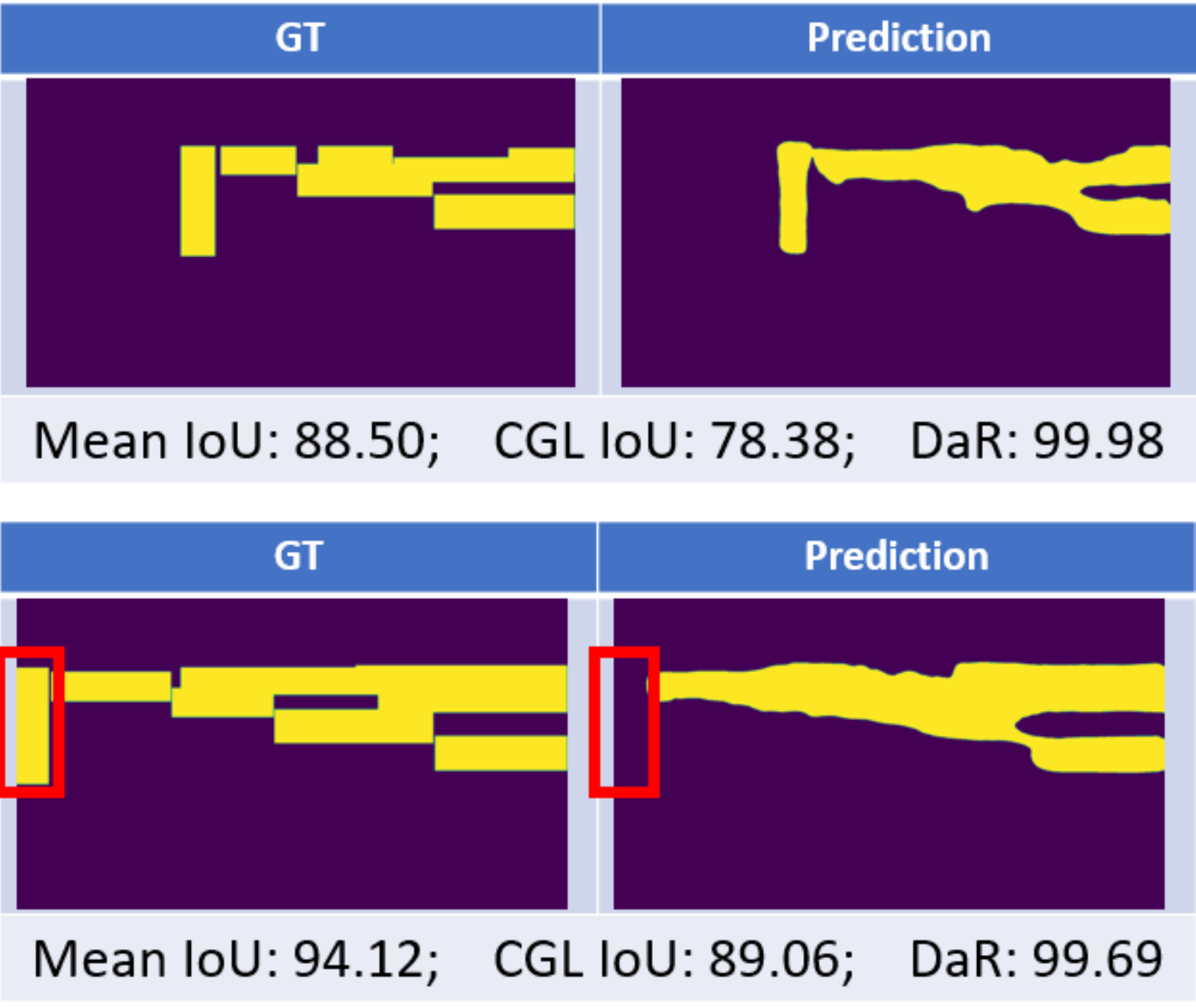}
  \caption{Different rows show GT and prediction for two different samples and the corresponding Mean IoU, CGL IoU, and DaR scores are mentioned at the bottom of each row. Yellow blobs indicate CGL blobs and the background/non-CGL class has been indicated using purple color. Red bounding boxes enclose regions where there is a difference between the two masks. This figure provides evidence of the fact that DaR scores are more apt for the evaluation of CGL segmentation models.}
  \label{eval_result_2}
\end{figure}
% \begin{figure}[h!]
%   \centering
%   \includegraphics[scale=0.4]{images/eval_result.pdf}
%   \caption{GT and predicted segmentation mask for two samples from the CGL dataset. The proposed model was used to get the segmentation masks. Different rows show GT and prediction for different samples and the corresponding Mean IoU, CGL IoU, and DaR scores are mentioned at the bottom of each row. Yellow blobs indicate CGL blobs and the background/non-CGL class has been indicated using purple color.}
%   \label{eval_result}
% \end{figure}

\begin{figure}[h!]
  \centering
  \includegraphics[scale=0.5]{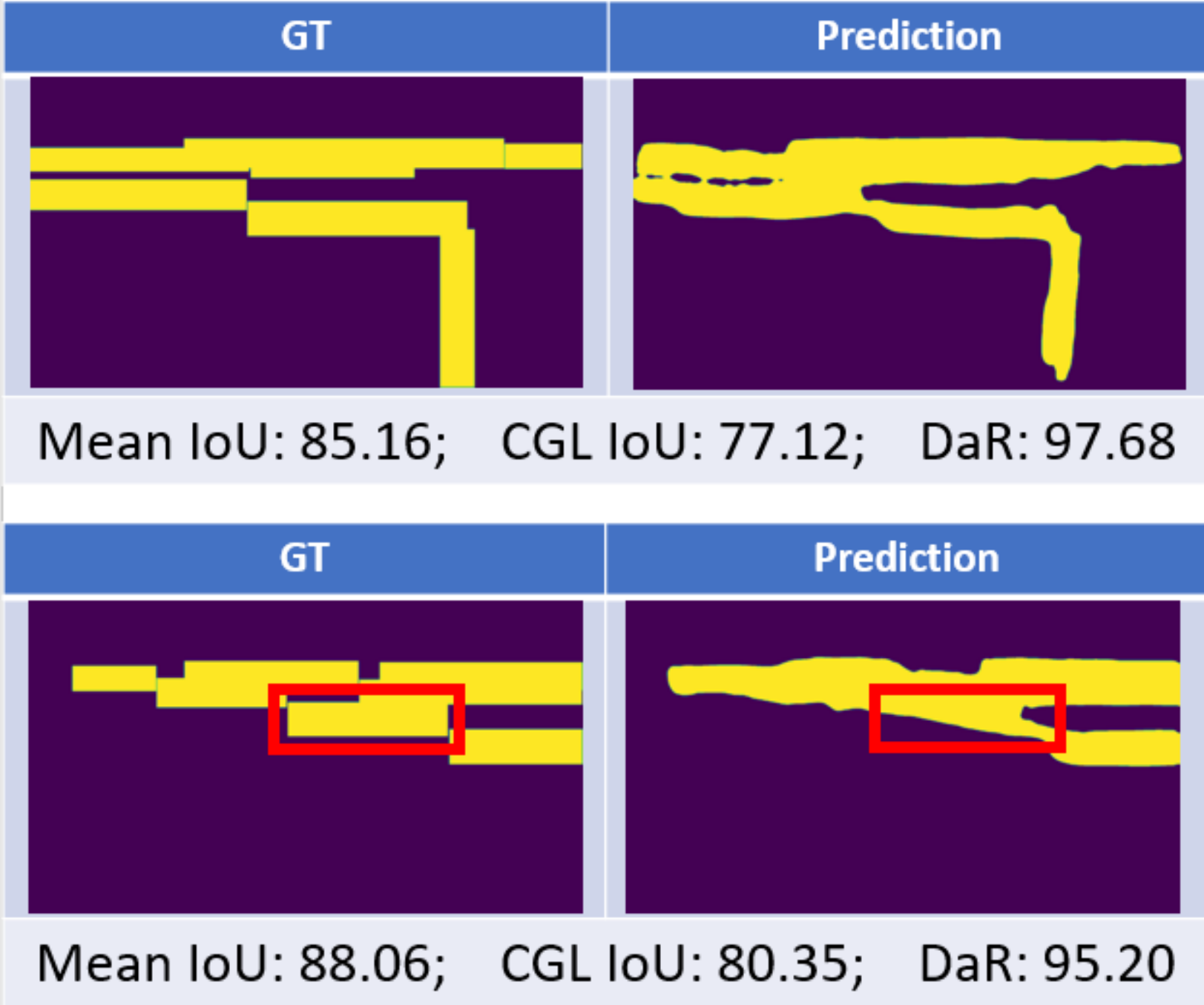}
  \caption{Yellow blobs indicate CGL blobs and the background/non-CGL class has been indicated using purple color. Red bounding boxes enclose regions where there is a difference between the two masks. DaR correctly gives more weightage to recognition than precise localization and thus provides more truthful scores to model outputs.}
  \label{eval_result_3}
\end{figure}

Finally, the output masks obtained after applying Gaussian blur are fused using element-wise OR operation to obtain y'. Then the output of the OR operation (y') is used along with the GT to compute the final score as follows:

\[
DaR = 1 - \frac{\text{Number of 1's in y'}}{\text{Number of 1's in GT}}
\]

Figure \ref{motivation_eval} shows two sample GTs and model predictions and corresponding metric scores (Mean IoU, CGL IoU). We can see that even though the prediction is very close to GT, IoU scores are low (scores should have been very close to 1), as in IoU score computation, matching is done at each and every coordinate in the prediction with the corresponding coordinate in the GT. Pixel level matching is not appropriate for CGL detection as the height and width of CGLs may not always be precisely defined. The proposed evaluation metric (DaR) overcomes this challenge and it correctly ignores minor deviations in height and width of predicted CGLs as compared to GT CGLs. As a result, the DaR score for the two samples justifiably increases to more than 99\%. Specifically, the DaR score for the sample shown in the first row is 99.38\% and that for the sample in the second row is 99.69\%.

Figures \ref{eval_result_2} \& \ref{eval_result_3} also show GT and the model prediction for a couple of image samples from CGL dataset. The output of the model matches with the GT in the sample shown in the first row, whereas the model has partially or completely missed (completely missed in figure \ref{eval_result_2} and partially missed in figure \ref{eval_result_3}) one CGL in the second scene (shown in the second row), indicated by overlaying red-colored bounding box over the two masks. However, the mean IoU and CGL IoU scores are higher for the predictions shown in the second row making them inapt for evaluation of CGL segmentation models. Our proposed model fairly evaluates the two predictions and correctly assigns a higher score to the model for the output shown in the first row. 

\newpage
\subsection{Qualitative Results}

\begin{figure*}[h!]
  \centering
  \includegraphics[scale=0.475]{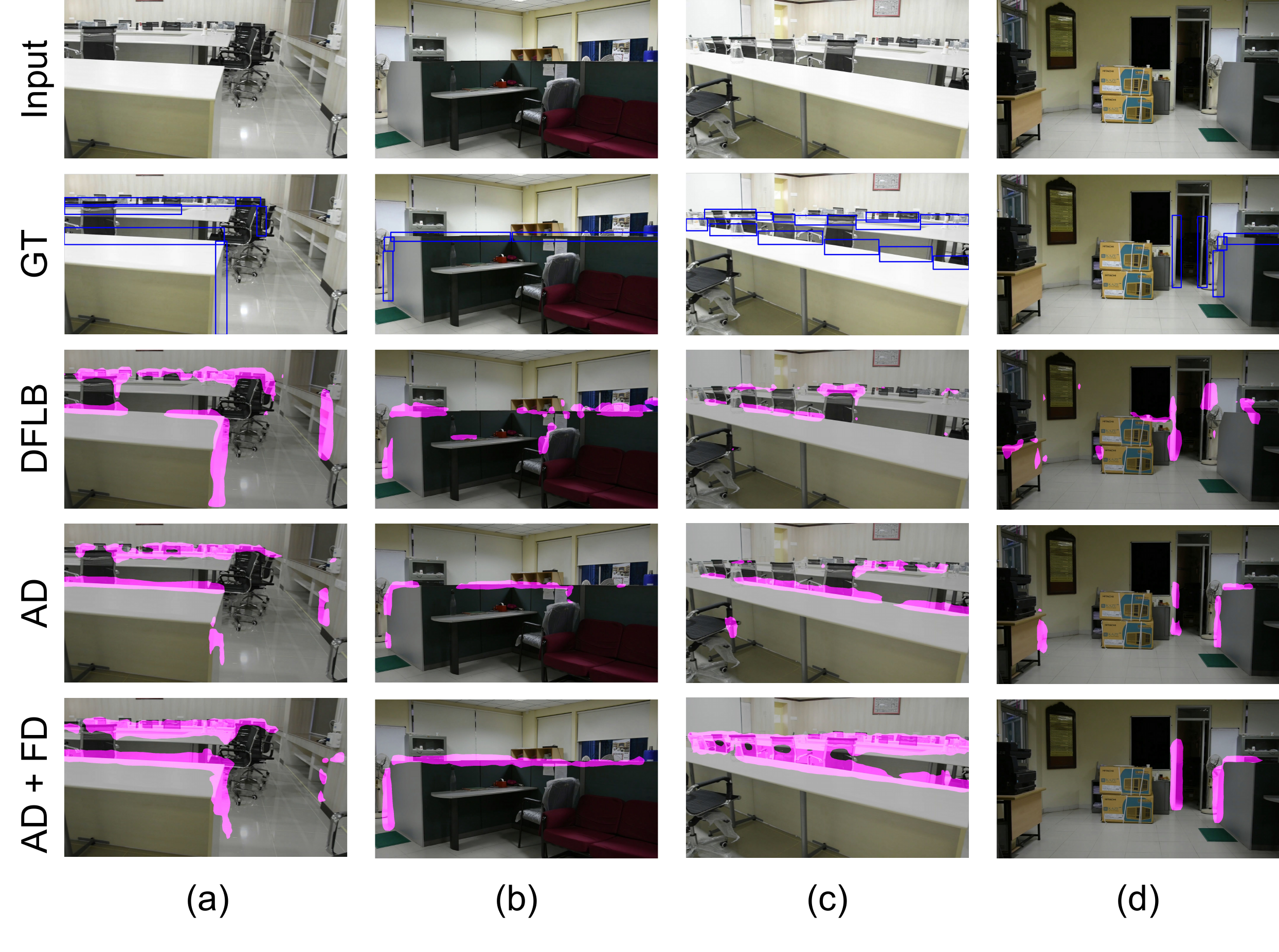}
  \caption{Qualitative results on split 1 of CGL dataset. Image regions classified as CGL by the models have been indicated by overlaying translucent pink-colored blobs over the image. The last two rows show the output of the model proposed in this paper [with HRNetv2 as the encoder]. The third row shows the output of the model proposed in \cite{saha2022catch} with the same encoder.}
  \label{paper2_qualitative_results}
\end{figure*}

Figure \ref{paper2_qualitative_results} shows qualitative results for the DFLB-based model proposed in \cite{saha2022catch} and the proposed method, on split 1 of the CGL dataset. DFLB-based model lacking semantic information fails to comprehend the spatial coverage of occluding items in the image and thus detects CGLs only partially (error in the spatial extent of detections) in most cases. Mere utilization of multitask learning with semantic segmentation as the auxiliary task alleviates the problem only slightly and we still see a lot of partial detections. However, when attention mechanism is employed as proposed in this paper, the problem of partial detection reduces significantly as evident from all the examples shown in figure \ref{paper2_qualitative_results}.

\newpage
\subsection{Quantitative Results}

As evident from tables \ref{results} \& \ref{results2}, the proposed CGL detection model outperforms the baseline and SOTA (\cite{saha2022catch}) model on CGL detection dataset \cite{saha2022catch}. Our proposed method effectively extracts and utilizes semantic class information to outperform other models on both train/test splits (irrespective of the encoder/decoder used). On split 1, we achieve a performance improvement of 3\% to 14\% mIoU and 3\% to 16\% DaR over SOTA and on split 2, performance improvement is $\approx$1\% mIoU and 1\% to 2\% DaR. For details about baseline models refer \cite{saha2022catch}.

\begin{table}[h!]
\centering
  \caption{Performance comparison of existing models with our proposed models on CGL detection. All models have been trained and tested using split 1. The first row in each section of the table reports the performance of the baseline model (i.e. when the proposed modules and the semantic head are not used) and the rest of the rows report performance of models specifically designed to perform CGL detection with the encoder and decoder architectures having the same base architecture as the ones in the baseline model. [*] indicates model was proposed in \cite{saha2022catch} and [\Cross] indicates model has been proposed in this paper.}
  \begin{center}
  \begin{tabular}{P{5.5cm}P{2cm}P{2cm}P{2cm}}
    \toprule
     Model & mIoU & CGL IoU & DaR \\
    \midrule
    MobileNetv2 + C1 \cite{sandler2018mobilenetv2} &  51.24 & 15.03 & 41.48 \\
    MobileNetv2 + C1 [*] &  54.19 & 20.62 & 44.78 \\
    MobileNetv2 + C1 (w/o attn) [\Cross] &  45.62 & 04.83 & 33.72 \\
    MobileNetv2 + C1 (w/ attn) [\Cross] &  \textbf{57.38} & \textbf{26.21} & \textbf{47.17} \\
    \midrule
    HRNetv2 + C1 \cite{sun2019high} &  55.31 & 23.36 & 44.16 \\
    HRNetv2 + C1 [*]  &  57.76 & 27.64 & 51.73 \\
    HRNetv2 + C1 (w/o attn)  [\Cross] &  66.27 & 42.25 & 59.61 \\
    HRNetv2 + C1 (w attn) [\Cross] & \textbf{67.55} & \textbf{44.74} & \textbf{61.23} \\
    \midrule
    ResNet + PPMDeepsup \cite{he2016deep} &  55.23 & 22.50 & 48.05 \\
    ResNet + PPMDeepsup [*] &  56.21 & 23.55 & 52.30 \\
    ResNet + PPMDeepsup (w/o attn) [\Cross] &  56.86 & 24.02 & 53.25 \\
    ResNet + PPMDeepsup (w attn) [\Cross] &  \textbf{70.46} & \textbf{49.58} & \textbf{68.26} \\
  \bottomrule
\end{tabular}
\end{center}
\label{results}
\end{table}

\begin{table}[h!]
\centering
  \caption{Performance comparison of existing models with our proposed models on CGL detection. This table provides evaluation scores for models on trained and tested using split 2. [*] indicates model proposed in \cite{saha2022catch} and [\Cross] indicates model proposed in this paper.}
  \begin{center}
  \begin{tabular}{P{5.5cm}P{2cm}P{2cm}P{2cm}}
    \toprule
     Model & mIoU & CGL IoU & DaR \\
    \midrule
    MobileNetv2 + C1 \cite{sandler2018mobilenetv2} &  76.72 & 60.42 & 76.84 \\
    MobileNetv2 + C1 + DFLB [*] &  78.46 & 63.40 & 80.32\\
    MobileNetv2 + C1 + AD [\Cross] & 66.57 & 43.42 & 60.04 \\
    MobileNetv2 + C1 + AD + FD [\Cross] & \textbf{78.47} & \textbf{63.42} & \textbf{80.51} \\
    \midrule
    HRNetv2 + C1 \cite{sun2019high} &  81.95 & 69.75 & 85.35 \\
    HRNetv2 + C1 + DFLB [*] &  83.55 & 72.38 & 87.64 \\
    HRNetv2 + C1 + AD [\Cross] & 83.76 & 73.53 & 87.95 \\
    HRNetv2 + C1 + AD + FD [\Cross] & \textbf{84.08} & \textbf{74.12} & \textbf{89.01} \\
    \midrule
    ResNet + PPMDeepsup \cite{he2016deep} & 80.60 & 67.30 & 83.27 \\
    ResNet + PPMDeepsup + DFLB [*] & 83.21 & 71.98 & 87.27 \\
    ResNet + PPMDeepsup + AD [\Cross] & 83.62 & 72.45 & 87.77 \\
    ResNet + PPMDeepsup + AD + FD [\Cross] & \textbf{83.98} & \textbf{73.19} & \textbf{88.84} \\
  \bottomrule
\end{tabular}
\end{center}
\label{results2}
\end{table}

\newpage
\newpage
\section{Conclusion}
This paper introduces a novel approach that can facilitate effective extraction and utilization of semantic class information, through the use of multi-task learning combined with self/cross-attention mechanism. The proposed method additionally helps to overcome the problem of unavailability of a large dataset for the training of deep CNN models. Semantic segmentation (task of the attention decoder) is used as the auxiliary task (along with the target task of CGL segmentation). Attention mechanism is used to propagate semantic class information to the fusion decoder which is trained to perform CGL segmentation. Consequently, we discuss the shortcomings of the standard IoU score and propose a better evaluation metric, named Dimension-agnostic Recall (DaR) for CGL detection. DaR gives more weightage to recognition than precise localization by ignoring minor differences in the height and width of predicted CGLs with respect to the ground truth CGL instances. 

% *****************************

% \cite{ref_article1,ref_book1,ref_proc1,ref_url1}.
\bibliographystyle{splncs04}
\bibliography{samplepaper}

\textbf{ACK: IMPRINT (MHRD/DRDO) GoI, for support}




% LNCS Homepage, \url{http://www.springer.com/lncs}. Last accessed 4
% \end{thebibliography}
\end{document}